\documentclass[conference]{IEEEtran}
\IEEEoverridecommandlockouts
\usepackage{booktabs}
\usepackage{multirow}
\usepackage{graphicx}
\usepackage{url}
\usepackage{CJKutf8}

\newcommand{\figref}[1]{Fig.\hspace{1mm}\ref{#1}}
\newcommand{\tabref}[1]{Table\hspace{1mm}\ref{#1}}

\newcommand{\squeezeup}{\vspace{-3.0mm}}

\begin{document}

\title{End-to-End Text Classification via Image-based Embedding using Character-level Networks}

\author{
  \IEEEauthorblockN{
    Shunsuke KITADA \hskip 2em Ryunosuke KOTANI \hskip 2em Hitoshi IYATOMI
  }
  \IEEEauthorblockA{
    \textit{Major in Applied Informatics, Graduate School of Science and Engineering} \\
    \textit{Hosei University}\\
    Tokyo, Japan \\
    \{shunsuke.kitada.8y@stu., ryunosuke.kotani.58@, iyatomi@\}hosei.ac.jp
  }
}

\maketitle

\begin{abstract}
 For analysing and/or understanding languages having no word boundaries based on morphological analysis such as Japanese, Chinese, and Thai,
 it is desirable to perform appropriate word segmentation before word embeddings.
 But it is inherently difficult in these languages.
 In recent years, various language models based on deep learning have made remarkable progress, and some of these methodologies utilizing character-level features have successfully avoided such a difficult problem.
 However, when a model is fed character-level features of the above languages, it often causes overfitting due to a large number of character types.
 In this paper, we propose a CE-CLCNN, character-level convolutional neural networks using a character encoder to tackle these problems.
 The proposed CE-CLCNN is an end-to-end learning model and has an image-based character encoder, i.e. the CE-CLCNN handles each character in the target document as an image.
 Through various experiments, we found and confirmed that our CE-CLCNN captured closely embedded features for visually and semantically similar characters and achieves state-of-the-art results on several open document classification tasks.
 In this paper we report the performance of our CE-CLCNN with the Wikipedia title estimation task and analyse the internal behaviour.
\end{abstract}


\section{Introduction}
Overfitting is one of the most essential problems in machine learning.
Various regularization methods have been proposed in order to improve generalization performance, especially in deep networks.
Data augmentation is the most common way to improve generalization performance of the system by increasing the training dataset in a pseudo manner.
In natural language processing (NLP) tasks, various data augmentation methods have also been proposed, such as synonym lists \cite{zhang2015character}, grammar induction \cite{jia2016data}, task-specific heuristic rules \cite{silfverberg2017data}, and contextual augmentation \cite{Kobayashi2018ContextualAD}.
However, these methods basically require appropriate word segmentation and semantic analysis of the context in advance,
and they are inherently difficult in the Asian languages, especially Japanese, Chinese or Thai, etc.
In recent years, various language models based on deep learning have made remarkable progress, and some of these methodologies utilizing character-level features have successfully avoided such problems \cite{zhang2015character, kim2016character}.
From the model selection point of view, recurrent neural networks (RNN) have been widely applied in NLP tasks, however, they have a significant problem in learning long text sequences.
The recent introduction of Long-short term memory (LSTM) \cite{hochreiter1997long} and gated recurrent units (GRU) \cite{chung2014empirical} alleviated this issue and are commonly used in the NLP field.
However, it still has drawbacks, such as difficulty in parallelization.
Character level convolutional neural networks (CLCNN) \cite{zhang2015character}, i.e. one-dimensional convolutional neural networks (CNN), also accept long text sequences, and, in addition, its training speed is generally faster than LSTM and GRU thanks to its native property and the ease of parallelization \cite{dauphin2016language, bradbury2016quasi}.
However, there are still problems remaining when dealing with the aforementioned languages.
When a model is fed character-level features (e.g. in one-hot vector or other common embeddings) of the language above, it often causes overfitting due to a large number of unique characters.
For example, Japanese\footnote{F. Sakade, Guide to Reading and Writing Japanese, 4th ed., J. Ikeda, Ed. Tuttle Publishing, 2013.} and Chinese\footnote{Table of General Standard Chinese Characters (\url{http://www.gov.cn/gzdt/att/att/site1/20130819/tygfhzb.pdf})} have over 2,000 type of characters in common use.
We need to tackle this problem as well.
Fortunately, a not insignificant number of Kanji and Han characters used in Japanese and Chinese are ideograms, which means its character shape represents its meaning.
Therefore, capturing the shape feature of characters in the document is meaningful for better understanding the contents.
Based on this hypothesis, several studies have been proposed recently. Shimada et al. \cite{shimada2016document} proposed epoch making schematics called image-based character embedding, in which they treat each character in the target document as an image.
Their model learns a low-dimensional character embedding by a convolutional auto-encoder (CAE) \cite{masci2011stacked} and then the relationship between the sequence of embeddings and the document category is trained with the following CLCNN.
They also proposed a simple and very effective data augmentation technique called wildcard training. The wildcard training randomly dropouts \cite{hinton2012improving} arbitrary elements in the embedded domain at the time of training of CLCNN.
This data augmentation method greatly improved system generalization performance without requiring morphological analysis.
They confirmed that the effect of this wildcard training improves the document classification accuracy by about 10\% on their evaluation, using open and private datasets.
Lately, Zhang et al. \cite{zhang2018word} also proposed similar semantic dropout for word representations and reported its effectiveness.
On the other hand, however, since their model learns CAE and CLCNN separately, it cannot fully exploit the merits of image-based character embedding.
Further improvement in performance can be expected.
Liu et al. \cite{liu2017learning} proposed an end-to-end document classification model that learns character embedding using a CNN-based character encoder and classifies documents using GRU on Chinese, Japanese, and Korean documents.
Unlike Shimada's model \cite{shimada2016document}, their model does not train to preserve the shape feature of characters explicitly.
But they demonstrated that characters with similar shape features are embedded closer to the character representation.
Su et al. \cite{DBLP:conf/emnlp/SuL17} proposed glyph-enhanced word embedding (GWE) to focus on the shape of Kanji.
The basic strategy is the same with \cite{shimada2016document}: GWE extracts the shape information of the character and uses it for training the word representation.
They also performed image-based character embedding on a Chinese document with CAE and showed that characters with similar shape features are represented by close character representation.
In these studies \cite{liu2017learning, DBLP:conf/emnlp/SuL17}, image-based character embedding showed promising performance, while there is still room for improvement from the viewpoint of introducing data augmentation in which the model inputs take advantage of the features of character image.
Based on these backgrounds, in this paper, we propose a new ``character encoder character-level convolutional neural networks'' (CLCNN) model.
The proposed CE-CLCNN is an end-to-end learning model and has an image-based character encoder.
Due to this architecture, our CE-CLCNN has the following desirable features:
\begin{enumerate}
  \item It is freed from intractable morphological analysis.
  \item It learns and obtains character embedding associating with character appearance.
  \item It is capable of a suitable data augmentation method both for image and embedded feature spaces.
\end{enumerate}
By introducing two essentially different types of data augmentation, the robustness of the model is enhanced and the performance of document classification task is significantly improved.

\begin{figure}[t]
 \centering
 \includegraphics[width=0.9\linewidth]{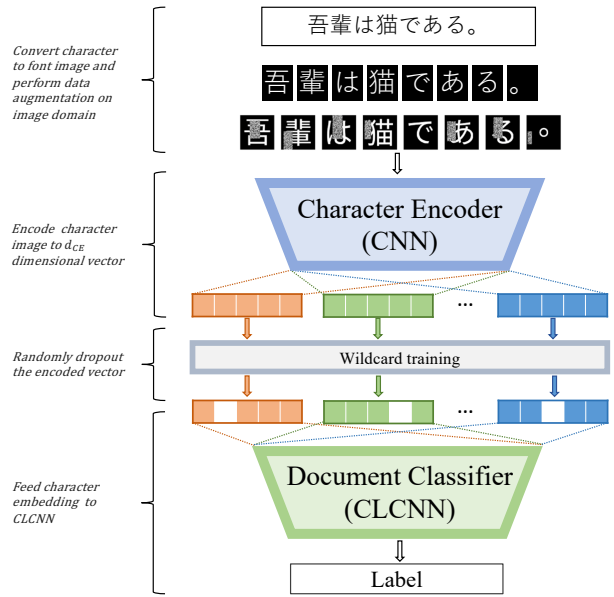}
  \caption{Schematics of our CE-CLCNN model. The CE-CLCNN is made up of character encoder (CE) component and document classification component. These two parts are basically composed of CNN and CLCNN respectively and are directly concatenated.}
  \label{fig:proposed_classification_system}
\end{figure}

\section{CE-CLCNN}
The outline of the proposed CE-CLCNN is shown in \figref{fig:proposed_classification_system}.
CE-CLCNN is made up of two different CNN consolidations.
The first CNN acts as a character encoder (CE) that learns character representations from character images, and the second CNN, CLCNN, performs document classification.
The parameters of these two consecutive networks are optimized by the backpropagation with the cross entropy error function as the objective function.

  \subsection{Character encoder by CNN}
  Firstly, each character of the target document is converted to an image having 36 $\times$ 36 pixels.
  The CE embeds (i.e. encodes) each character image into a $d_{CE}$ dimensional feature vector.
  \tabref{tab:architecture_of_character_encoder} shows the architecture of CE used in this instance.
  Here, let $k$ be the kernel size and $o$ be the number of filters.
  In the training of CE, continuous $C$ characters in the document are treated as a chunk.
  The convolution is performed with depth-wise manner, and, in fact, each input character is embedded in 128 dimensional vector.
  Thus, the input dimension of the CE is 36$\times$36$\times C$, and the output of that is 1$\times$128$\times C$.
  The CE trains with the batch size of $\mathcal{B}$.

  \begin{table}[t]
   \squeezeup
   \caption{Architecture of Character Encoder (CE)}
   \label{tab:architecture_of_character_encoder}
   \centering
   \begin{tabular}{cc}
    \toprule
    Layer \#  & CE configuration \\
    \midrule
    1 & Conv($k$=(3, 3), $o$=32) $\rightarrow$ ReLU \\ \addlinespace
    2 & Maxpool($k$=(2, 2)) \\ \addlinespace
    3 & Conv($k$=(3, 3), $o$=32) $\rightarrow$ ReLU \\ \addlinespace
    4 & Maxpool($k$=(2, 2)) \\ \addlinespace
    5 & Conv($k$=(3, 3), $o$=32) $\rightarrow$ ReLU \\ \addlinespace
    6 & Linear(800, 128) $\rightarrow$ ReLU \\ \addlinespace
    7 & Linear(128, 128) $\rightarrow$ ReLU \\
    \bottomrule
   \end{tabular}
  \end{table}

  \subsection{Document classifier by CLCNN}
  The character representation of the $d_{CE}$ bit/character encoded from the CE is reshaped to be the batch size $\mathcal{B}$ again with the character string length of $\mathcal{C}$.
  Then the representations are inputted to the CLCNN.
  Note that we use convolutions with stride $s$ rather than pooling operations which are widely used in natural language processing, with reference to prior work \cite{zhang2017deconvolutional}.
  \tabref{tab:architecture_of_clcnn} shows the architecture of CLCNN used in this instance.

  \begin{table}[t]
   \squeezeup
   \caption{Architecture of Character-level Convolutional Neural Network (CLCNN)}
   \label{tab:architecture_of_clcnn}
   \centering
   \begin{tabular}{cc}
    \toprule
    Layer \# & CLCNN configuration \\
    \midrule
    1 & Conv($k$=(1, 3), $o$=512, $s$=3) $\rightarrow$ ReLU \\ \addlinespace
    2 & Conv($k$=(1, 3), $o$=512, $s$=3) $\rightarrow$ ReLU \\ \addlinespace
    3 & Conv($k$=(1, 3), $o$=512) $\rightarrow$ ReLU \\ \addlinespace
    4 & Conv($k$=(1, 3), $o$=512) \\ \addlinespace
    5 & Linear(5120, 1024) \\ \addlinespace
    6 & Linear(1024, \# classes) \\
    \bottomrule
   \end{tabular}
  \end{table}

  \subsection{Data augmentation on input space and feature space}
  \begin{figure}[t]
   \centering
   \includegraphics[width=\linewidth]{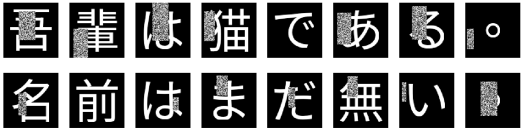}
   \caption{Example of data augmentation on image domain (random erasing data augmentation \cite{zhong2017random}). Note that this is an example of an implementation on this experiment and augmentation is not limited to this method.}
   \label{fig:visualize_random_erasing_chars}
  \end{figure}
  Convolutional neural networks are known to require a large amount of diverse training data.
  Our CE-CLCNN model has a capability to perform data augmentation both in the input space and the feature space thanks to its end-to-end structure.
  In the input space, we apply random erasing data augmentation (RE) \cite{zhong2017random} to the character image that will be fed to the CE.
  Each character image is randomly masked with noise on the rectangular area, and thus a part of the character is occluded as shown in \figref{fig:visualize_random_erasing_chars}.
  In the feature embedded space, we apply wildcard training (WT) \cite{shimada2016document} that randomly drops out some of embedded expression
  (i.e. some element of encoded vector) with the ratio of $\gamma_w$.
  
  \begin{table}[htbp]
   \squeezeup
   \caption{Parameters of Random erasing data augmentation}
   \label{tab:parameters_of_random_erasing_data_augmentation}
   \centering
    \begin{tabular}[tb]{lc}
     \toprule
     Parameter & Scale \\
     \midrule
     Erasing probability $p$ & 0.3  \\ \addlinespace
     Max area ratio $s_l$    & 0.4  \\ \addlinespace
     Min area ratio $s_h$    & 0.02 \\ \addlinespace
     Max aspect ratio $r_1$  & 2.0  \\ \addlinespace
     Min aspect ratio $r_2$  & 0.3  \\
    \bottomrule
    \end{tabular}
  \end{table}

\section{Experiments}
   \subsection{Implementation}
   The number of embedding dimensions and the chunk size of characters were set to $d_{CE} = 128$ and $C = 10$, respectively.
   \tabref{tab:parameters_of_random_erasing_data_augmentation} summarizes the parameters used in random erasing data augmentation.
   The ratio in the wildcard training was set to $\gamma = 0.1$.
   In the CLCNN, the batch size of the embedded characters in the training $B = 256$, and Adam \cite{kingma2014adam} was used for parameter optimization. 
   
   \subsection{Category estimation of Wikipedia titles}
   In this paper, we evaluate our proposed CE-CLCNN using an open dataset for category estimation of Wikipedia titles.
   The Wikipedia title dataset \cite{liu2017learning} contains the article titles acquired from Wikipedia and the related topic class label.
   This dataset includes 12 classes: Geography, Sports, Arts, Military, Economics, Transportation, Health Science, Education, Food Culture, Religion and Belief, Agriculture and Electronics.
   In this experiment, we used the Japanese data subset this time (total 206,313 titles).
   For training of the model, we split the dataset into the training and testing set with an 8:2 ratio, respectively.
   Zero padding was performed for titles with less than 10 characters so that the input sentence would be 10 characters or more.
   \tabref{tab:results_of_wikipedia_titles_ja} shows the results. To the best of our knowledge, the proposed CE-CLCNN showed state-of-the art performance on this dataset.
   Shimada's method \cite{shimada2016document} showed about 4\% better performance than Liu's method \cite{liu2017learning} (proposed later thanks to their WT with highly effective generalization).
   The proposed CE-CLCNN with RE and WT showed even better performance by about 4\%.
   According to \tabref{tab:results_of_wikipedia_titles_ja}, the performance of the native CE-CLCNN (i.e. without RE and WT) was equivalent to Shimada's CLCNN+WT.
   Since the performance improvement of CE-CLCNN by introduction of WT was limited, we can speculate CE-CLCNN has sufficient model versatility in the embedded space.
   While on the other hand, the effect of introducing RE was certain, with a 3-3.5\% gain.

   \begin{table}[htbp]
    \caption{Results of category estimation of Wikipedia titles}
    \label{tab:results_of_wikipedia_titles_ja}
    \centering
    \begin{tabular}{lr}
     \toprule
     Method                                            & Accuracy[\%] \\
     \cmidrule(lr){1-1} \cmidrule(lr){2-2}
     \textbf{(Proposed)} RE + CE-CLCNN + WT            & \textbf{58.4} \\[2pt]
     \textbf{(Proposed)} RE + CE-CLCNN                 & 58.0 \\[2pt]
     \textbf{(Proposed)} CE-CLCNN + WT                 & 55.3 \\[2pt]
     \textbf{(Proposed)} CE-CLCNN                      & 54.4 \\[2pt]
     CLCNN + WT$\dagger$ \cite{shimada2016document}    & 54.7 \\[2pt]
     CLCNN$\dagger$ \cite{shimada2016document}         & 36.2 \\[2pt]
     VISUAL model$\ddag$ \cite{liu2017learning}               & 47.8 \\[2pt]
     LOOKUP model$\ddag$ \cite{liu2017learning}               & 49.1 \\[2pt]
     Ensemble (VISUAL + LOOKUP)$\ddag$ \cite{liu2017learning} & 50.3 \\[2pt]
     \bottomrule
     \multicolumn{2}{l}{$\dagger$ Not published} \\
     \multicolumn{2}{l}{$\ddag$ Refering to Liu et. al. \cite{liu2017learning}}
    \end{tabular}
   \end{table}

   \subsection{Analysis of Character Encoder}
   \begin{figure*}[t]
   \centering
     \begin{tabular}[tb]{c}
      \begin{minipage}{0.5\hsize}
       \centering
       \includegraphics[width=\linewidth]{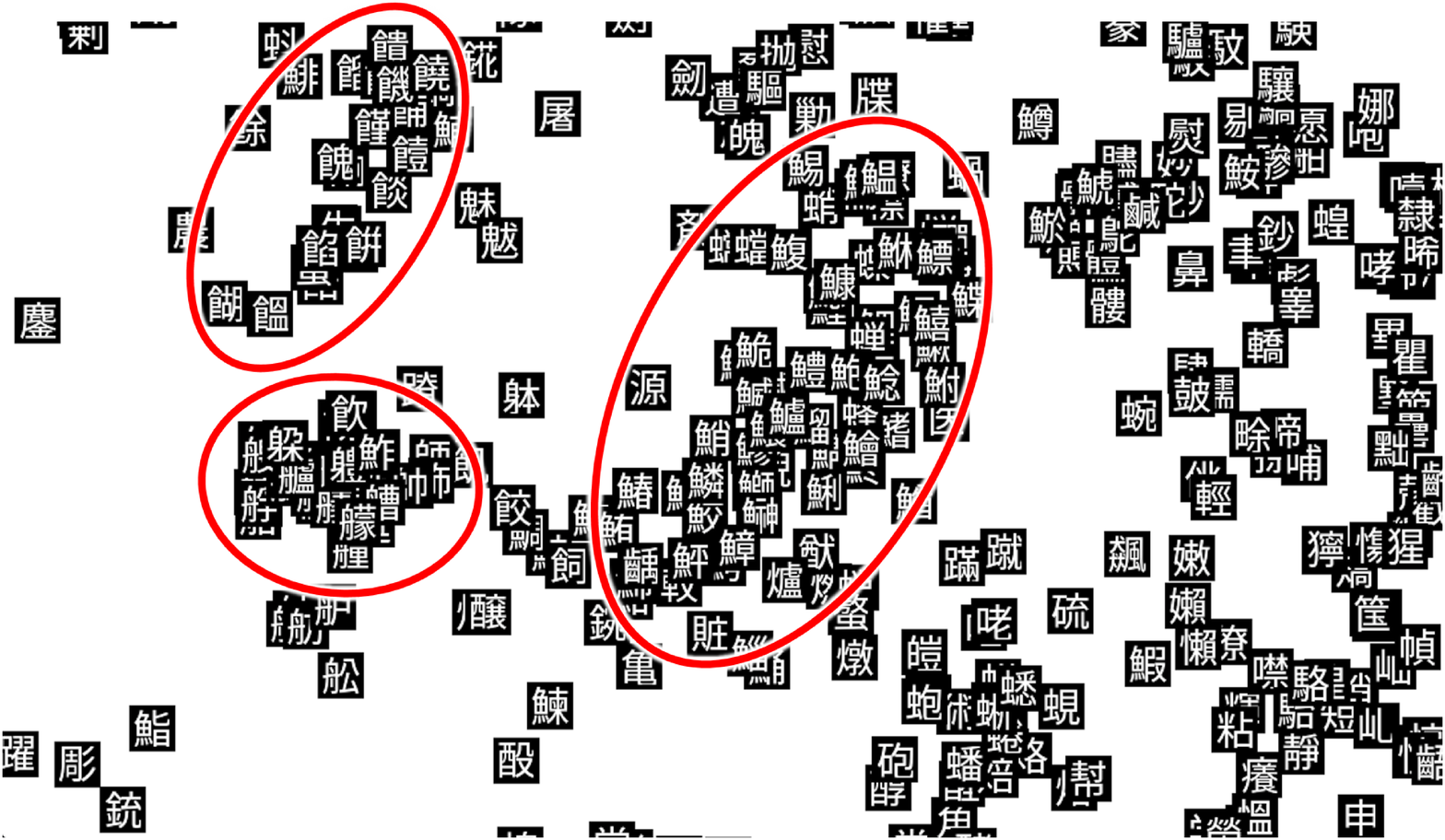}
      \end{minipage}

      \begin{minipage}{0.5\hsize}
       \centering
       \includegraphics[width=\linewidth]{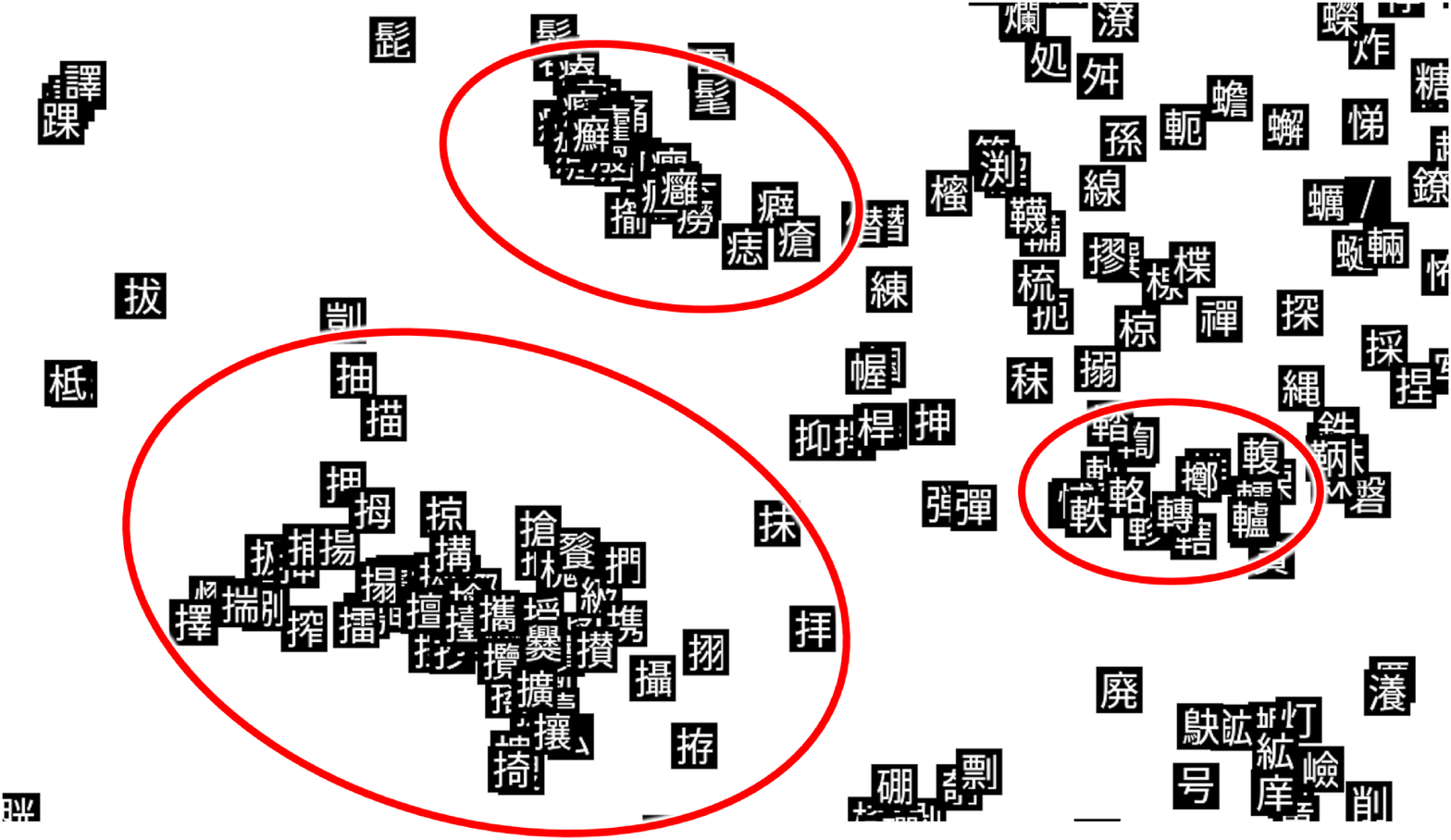}
      \end{minipage}

     \end{tabular}
    \caption{Example of character representation in the feature embedded domain obtained by our CE-CLCNN. The 128 dimensional feature space was mapped into 2 dimension with t-SNE \cite{maaten2008visualizing} for visualization purpose. We can see that similarly shaped and semantically close characters are located near one another.}
    \label{fig:example_of_t-sne_visualization}
   \end{figure*}

   \begin{CJK*}{UTF8}{ipxm}
   \begin{table}[h]
    \squeezeup
    \caption{Top five characters near the character representation for the query character}
    \label{tab:result_character_encoder}
    \centering
    \begin{tabular}{ccc}
     \toprule
     Query character       &  Neighbouring character & \begin{tabular}{c}
						      Euclidean distance \\
						     with query character \end{tabular} \\
     \midrule
     \multirow{5}{*}{{\LARGE 鮫}} &  鰭     & 370.1 \\
                                  &  駮     & 403.7 \\
                                  &  鮪     & 405.2 \\
                                  &  鰐     & 409.4 \\
                                  &  鰤     & 409.6 \\
     \midrule \midrule
     \multirow{5}{*}{{\LARGE 痛}} &  癨     & 317.2 \\
                                  &  癜     & 388.3 \\
                                  &  瘻     & 398.3 \\
                                  &  痕     & 398.9 \\
                                  &  痴     & 399.2 \\
     \midrule \midrule
     \multirow{5}{*}{{\LARGE 披}} &  彼     & 452.8 \\
                                  &  擅     & 491.5 \\
                                  &  擔     & 520.5 \\
                                  &  擒     & 533.8 \\
                                  &  捗     & 536.8 \\
    \bottomrule
    \end{tabular}
   \end{table}
   \end{CJK*}
   
   \tabref{tab:result_character_encoder} shows an example  of similar characters in the CE-encoded feature embedding domain with the 5-nearest neighbouring method.
   Many of the neighbouring characters for the query character were similar in shape features of letters, such as radicals (i.e. character components).
   Therefore, it was confirmed that the character encoder learned by capturing the shape feature of the character.

   \begin{CJK*}{UTF8}{ipxm}
   Furthermore, we extracted the character representation of Chinese characters by using learned CE, and then projected the representation on 2 dimensional space using t-SNE \cite{maaten2008visualizing}.
   A part of the visualization result is shown in \figref{fig:example_of_t-sne_visualization}.
   We can see that characters with the same components are clustered.
   Note that Su et. al. \cite{DBLP:conf/emnlp/SuL17} explicitly learned to preserve character shape features by CAE,
   but our CE-CLCNN does not explicitly learn character representation that preserves the shape features of characters explicitly.
   In CE-CLCNN, since the loss of document classification backpropagates to the CE which learns character representation,
   we found that clusters that are semantically similar in character representation are close clusters.
   For example, it can be seen that the character cluster having ``舟'' component representing ``boat'' and the character cluster having ``魚'' component representing ``fish'' are closely related. 
   \end{CJK*}

\section{Conclusion}
In this paper, we propose the new and promising text analysis model ``CE-CLCNN'' to solve several conventional problems for languages such as Japanese and Chinese.
We confirm not only its excellent document classification performance, but also its readability in terms of how the model works.
In near future, we would like to investigate our model more and apply it to other languages whose character shapes are related to the meaning.

{\scriptsize
  \bibliographystyle{IEEEtran}
  \bibliography{reference}
}

\end{document}